\documentclass{article}

\usepackage{arxiv}
\usepackage{makecell}

\usepackage[utf8]{inputenc} 
\usepackage[T1]{fontenc}    
\usepackage{url}            
\usepackage{booktabs}       
\usepackage{amsfonts}       
\usepackage{nicefrac}       
\usepackage{microtype}      
\usepackage{lipsum}
\usepackage{graphicx}
\usepackage[sort]{natbib} 
\usepackage{xcolor}
\usepackage[colorlinks = true,
            linkcolor = blue,
            urlcolor  = blue,
            citecolor = blue,
            anchorcolor = blue]{hyperref}
\newcommand{\MYhref}[3][blue]{\href{#2}{\color{#1}{#3}}}
\usepackage{latexsym}
\usepackage[nocompress]{cite}
\usepackage{cite}

\usepackage[labelfont=bf]{caption}

\usepackage{subcaption}

\title{COFGA: A Dataset for Fine-Grained Classification of Objects from Aerial Imagery}

\author{Eran Dahan \\
        \and
        Tzvi Diskin \\
        \and
        Amit Amram\\
        \and
        Amit Moryossef\\
        \and
        Omer Koren}

\begin{document}
\maketitle
\begin{abstract}
Detection and classification of objects in overhead images are two important and challenging problems in computer vision. Among various research areas in this domain, the task of fine-grained classification of objects in overhead images has become ubiquitous in diverse real-world applications, due to recent advances in high-resolution satellite and airborne imaging systems. The small inter-class variations and the large intra-class variations caused by the fine-grained nature makes it a challenging task, especially in low-resource cases. In this paper, we introduce COFGA—a new open dataset for the advancement of fine-grained classification research. The 2,104 images in the dataset are collected from an airborne imaging system at 5–15 cm ground sampling distance (GSD), providing higher spatial resolution than most public overhead imagery datasets. The 14,256 annotated objects in the dataset were classified into 2 classes, 15 subclasses, 14 unique features, and 8 perceived colors—a total of 37 distinct labels—making it suitable to the task of fine-grained classification more than any other publicly available overhead imagery dataset. We compare COFGA to other overhead imagery datasets and then describe some distinguished fine-grain classification approaches that were explored during an open data-science competition we have conducted for this task (\MYhref{https://mafatchallenge.mod.gov.il/}{The MAFAT Challenge}). 
\end{abstract}

\keywords{Dataset \and Aerial imagery \and Computer vision \and Fine-grained classification \and Multilabel learning algorithms \and Ensemble methods}

\section{Introduction}\label{sec:introduction}
Classification techniques and the amount of footage captured by aerial sensors have been growing rapidly in recent decades. An abundance of images leads to an overflow of information that cannot be processed by human analysts alone; therefore, the vast majority of aerial imagery is unlabeled. The emerging technologies of machine learning and artificial intelligence have begun to address this challenge, and although some progress has been made to solve this task \citep{objectrecognision,rotationinvarient}, there are still problems that need to be addressed. One of the challenges is the lack of large enough annotated datasets required to train deep neural networks to automatically classify the different objects as seen from the air; another challenge is the research required for a neural network architecture that is suitable for low-resource training; another challenge is to differentiate between hard-to-distinguish object classes, i.e., fine-grained classification. A variety of techniques were developed
to address these challenges. \citep{attention} use an attention mechanism in order to select relevant patches to a certain object for dealing with the small inter-class variations,
while \citep{noise} propose the use of large-scale noisy data in order to significantly decrease the cost of annotating the dataset; finally, the research required to build a representation for the different features that can be jointly learned for different classes is also a challenge. Solving these challenges can give rise to different applications in the domain of overhead imagery analysis and exploitation. 

In the fall of 2018, Israel Ministry Of Defense's Directorate of Defense Research and Development (IMOD’s DDR\&D), aka MAFAT, released COFGA, a dataset for the detection and classification of fine-grained features of vehicles in high-resolution aerial images, along with a public prize competition named the “MAFAT Challenge”.

In this paper, we describe the COFGA dataset (\S\ref{sec:cofga}), compare it to related leading datasets  (\S\ref{sec:related}\& \S\ref{sec:comparison}) 
and detail different approaches to tackle the MAFAT challenge as discovered during the competition (\S\ref{sec:challange}). Particularly, we describe the in-depth approaches of two notable solutions, as well as the baseline solution based on existing state-of-the-art models
(\S\ref{sec:part}).

\begin{figure}
    \centering
    \hspace{2cm}
    \begin{subfigure}[b]{\textwidth}
    \includegraphics[width=\textwidth]{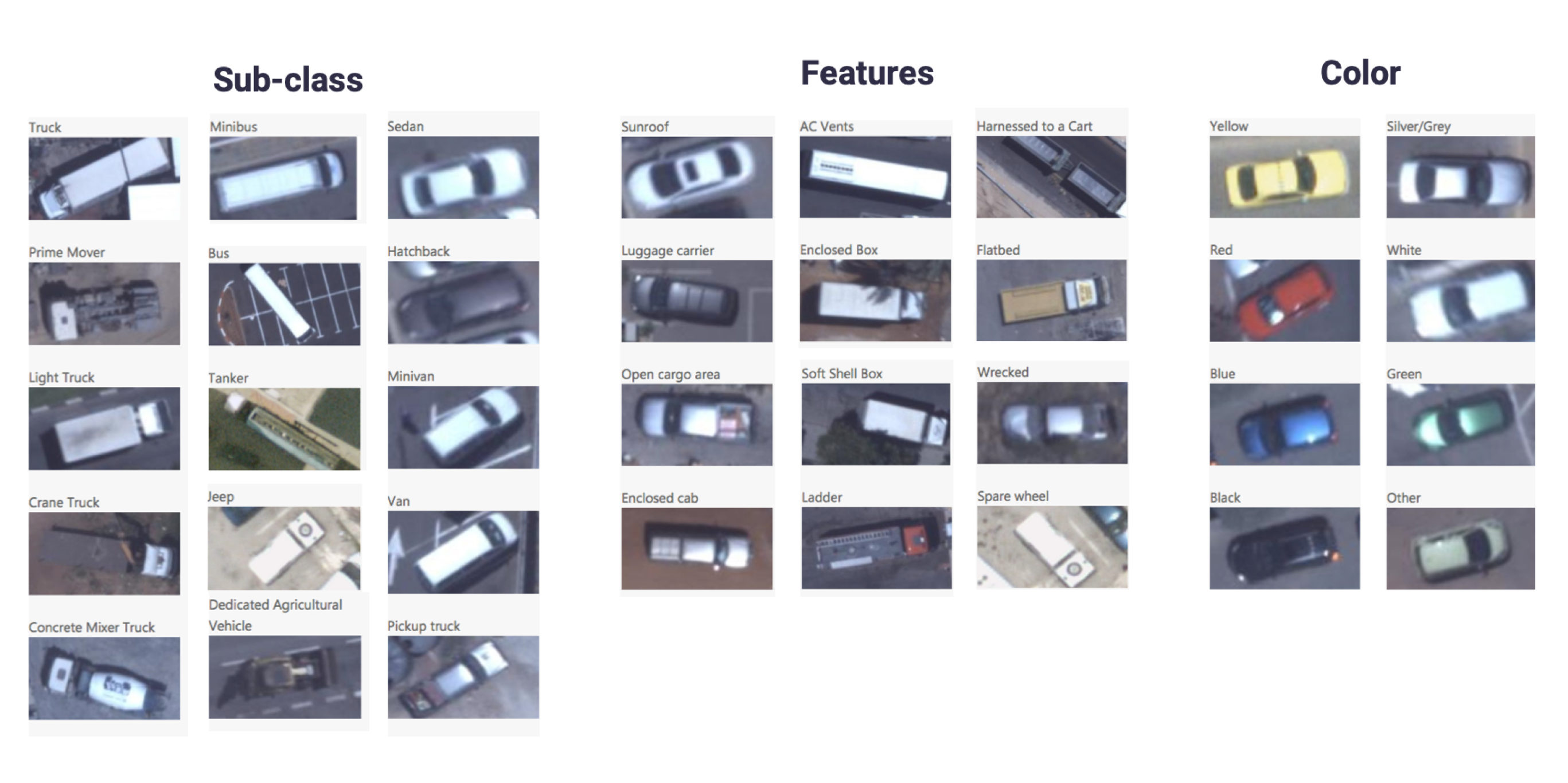}
    \end{subfigure}
    \hspace{5cm}
    \hspace{3cm}
    \begin{subfigure}[b]{0.24\textwidth}
    \caption{Subclasses}
    \label{fig:challenge-labels:subclass}
    \end{subfigure}
    \hspace{1.3cm}
    \begin{subfigure}[b]{0.24\textwidth}
    \caption{Unique features}
    \label{fig:challenge-labels:features}
    \end{subfigure}
  \hspace{1.3cm}
  \begin{subfigure}[b]{0.24\textwidth}
    \caption{Perceived color}
    \label{fig:challenge-labels:color}
    \end{subfigure}\\

\caption{A sample of COFGA's fine-grained classification labels, including subclasses, unique features and perceived color}
\label{fig:dist}
\end{figure}

\section{Related Work}\label{sec:related}
While searching datasets for training deep neural networks for the task of fine-grained classification from overhead images, one can find that there are several public datasets of labeled aerial and satellite imagery. In this section, we describe notable such datasets that include fine-grained features of objects in overhead images.

\subsection{xView—Objects in context in overhead imagery} This dataset \citep{xview} contains 1,127 satellite images with spatial resolution of 30 cm GSD, each image is in a size of about 4,000 × 4,000 pixels. xView has ˜1M annotated objects. The dataset's ontology includes two granularity levels (parent and child-level classes), and it has 7 main classes, each containing 2–13 subclasses for a total of 60 different subclasses. It contains ˜280K instances of vehicles. xView uses horizontal, axis-aligned, bounding boxes (BB) annotation method. Each BB is represented by 4 parameters and contains redundant pixels—pixels that don't belong to the actual object but do fall inside the BB. This creates a large variance between two different samples of the same object. Additionally, in crowded scenes, two or more neighboring BB can overlay each other (Fig \ref{fig:bb}b), which makes classification more difficult.
\newline
xView is used for object detection and classification.

\subsection{DOTA-v1.5—Dataset for Object deTection in Aerial images}
This dataset \citep{dota} contains 2,806 satellite images from multiple sensors and platforms (e.g. Google Earth) with multiple resolutions. The typical spatial resolution of images in this dataset is 15 cm GSD. Each image is in a size of about 4,000 × 4,000 pixels. DOTA-v1.5 contains ˜470K annotated object instances, each of which is assigned with one of 16 different classes. This dataset contains 380K vehicle instances divided among 2 categories—‘small vehicle’ and ‘large vehicle’. The dataset's ontology includes a single granularity level (no subclasses). Objects are annotated with the oriented BB (Fig \ref{fig:bb}c) method. Each BB is represented by 8 parameters. Compared to horizontal BB, oriented BB reduce the number of redundant pixels surrounding an object and reduce the overlap area between neighboring BB. 
\newline
DOTA-v1.5 is used for object detection and classification.

\subsection{iSAID—Instance Segmentation in Aerial Images Dataset} 
This dataset \citep{isaid} is built on the DOTA dataset and contains the same 2,806 satellite images. The difference is the annotation method; unlike DOTA, iSAID uses polygon segmentation. Each object instance is independently annotated from scratch (not using DOTA's annotations) and is represented by the exact coordinates of the pixels surrounding the object. Polygon segmentation (Fig \ref{fig:bb}d) minimizes the number of redundant pixels and removes the overlap area of neighboring object crops, which makes it more reliable for accurate detection, classification and segmentation. 
\newline
iSAID is mainly used for pixel-level segmentation and object separation but can also be used for object detection and classification.

\subsection{COWC—Cars Overhead With Context}
\begin{figure}
\raggedright
\begin{subfigure}{\linewidth}
\centering
    \includegraphics[width=\linewidth]{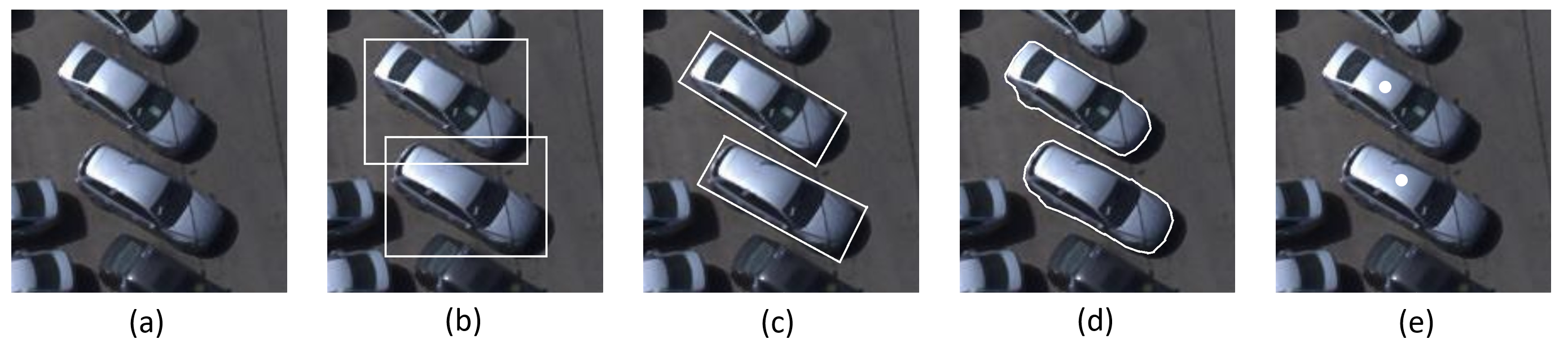}
\end{subfigure}\\
\caption{Visualization of different annotation methods: (a) an image patch, (b) horizontal, axis-aligned, BB, (c) oriented BB, (d) polygon segmentation, (e) CPM}
\label{fig:bb}
\end{figure}

This dataset \citep{cowc} contains 2,418 overhead images from six different sources. The images were standardized to 15 cm GSD and to a size of 1,024 × 1,024 pixels. COWC contains 32,716 car instances and 58,247 negative examples. The COWC dataset's granularity level is 1 and it is annotated only with the "centroid pixel map" (CPM) of each object (Fig \ref{fig:bb}e). This annotation method is very easy and rapid, but it mostly allows object counting as its usability for detection and classification tasks is limited. 

\subsection{VisDrone}
This dataset \citep{visdrone} contains 400 videos and 10,209 images, captured by various drone-mounted cameras. Each image is in a size of 1,500 × 1,500 pixels. Images, which are of high spatial resolution, are captured from various shooting angles (vertical and oblique). VisDrone contains 2.5M object instances that are divided among 10 different categories. This dataset contains ~300K vehicle instances divided among 6 categories. The VisDrone dataset's granularity level is 1 and it is annotated with the horizontal BB method.
\raggedright\newline

VisDrone is used for object detection, classification, tracking and counting.\newline

\section{The COFGA Dataset}\label{sec:cofga}

The dataset we present here is an extensive and high-quality resource that will hopefully enable the development of new and more accurate algorithms. Compared to other aerial open datasets, it has two notable advantages. First, its spatial resolution is very high (5–15 cm GSD). Second, and most prominent, is that the objects are tagged with fine-grained classifications referring to the delicate and specific characteristics of vehicles, such as air conditioning (AC) vents, the presence of a spare wheel, a sunroof, etc (Fig \ref{fig:challenge-labels:features}).

\subsection{Dataset Details}
\begin{figure}
\centering
\begin{subfigure}{.3\textwidth}
  \centering
  \includegraphics[width=.64\linewidth,height=50mm]{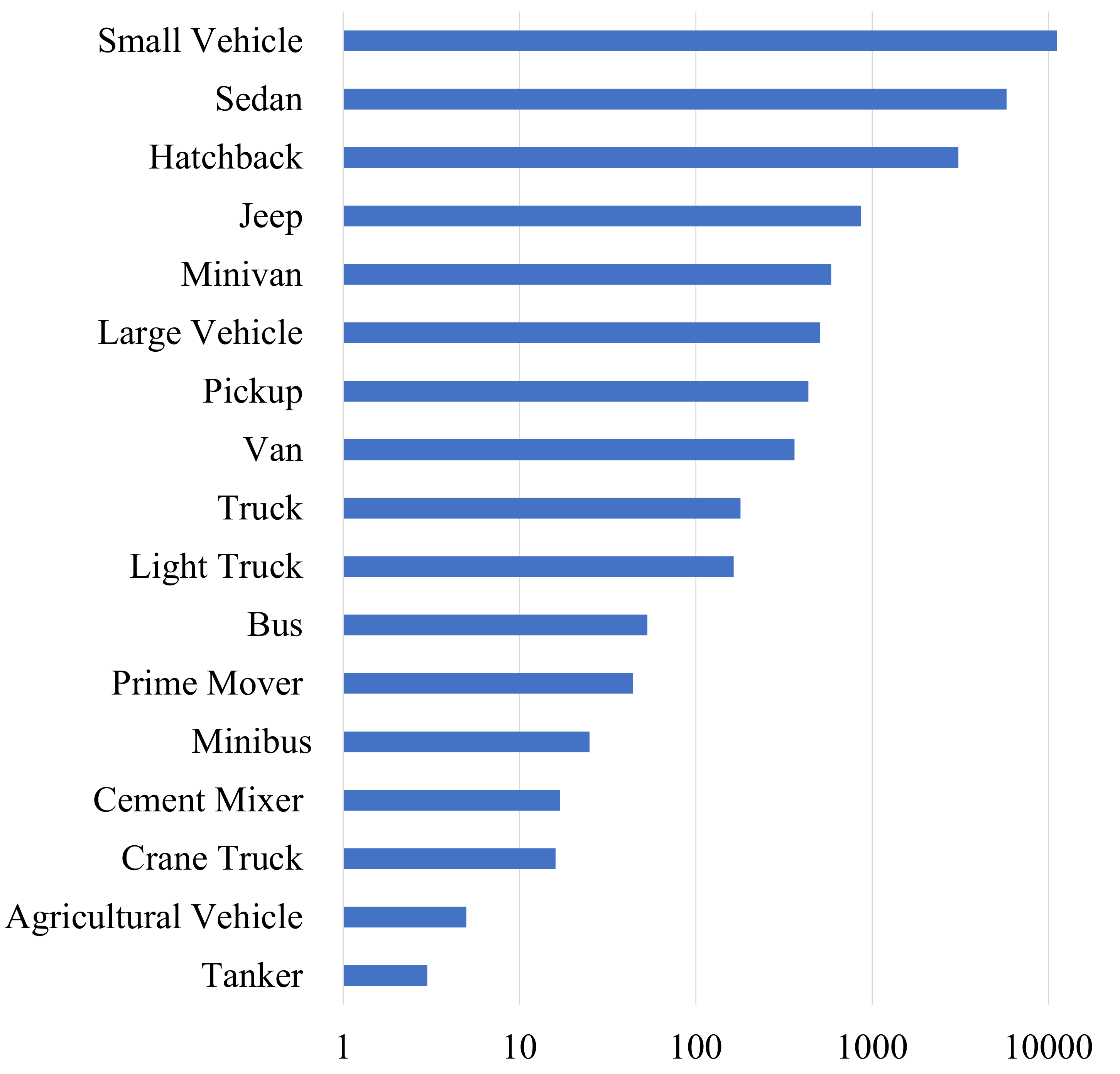}
  \caption{Number of items in vehicle subclasses}
  \label{fig:vehicles}
\end{subfigure}%
\begin{subfigure}{.64\textwidth}
  \centering
  \includegraphics[width=.64\linewidth,height=50mm]{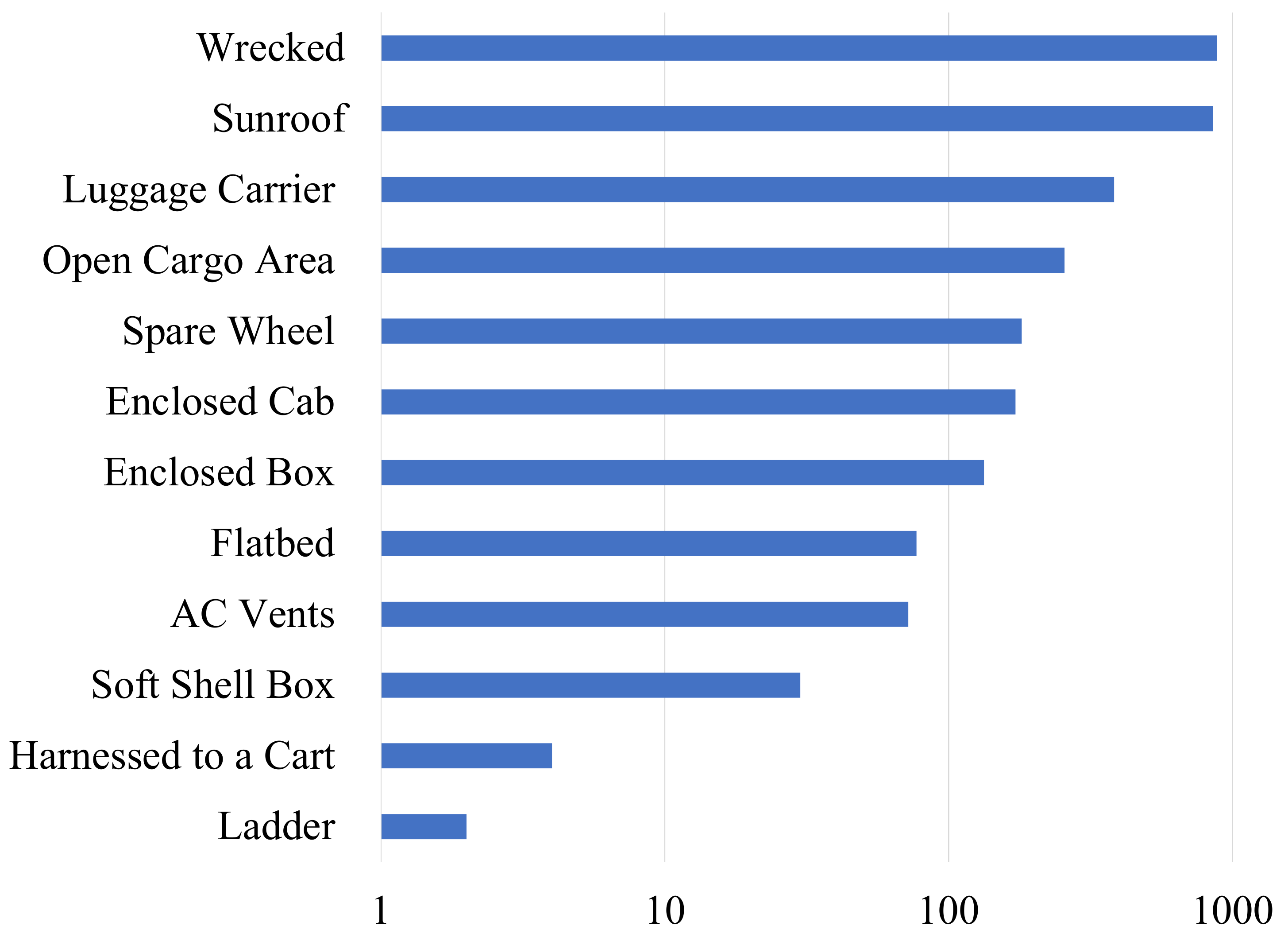}
  \caption{Number of items in feature classes}
  \label{fig:features}
\end{subfigure}
\begin{subfigure}{.64\textwidth}
  \centering
  \includegraphics[width=.64\linewidth]{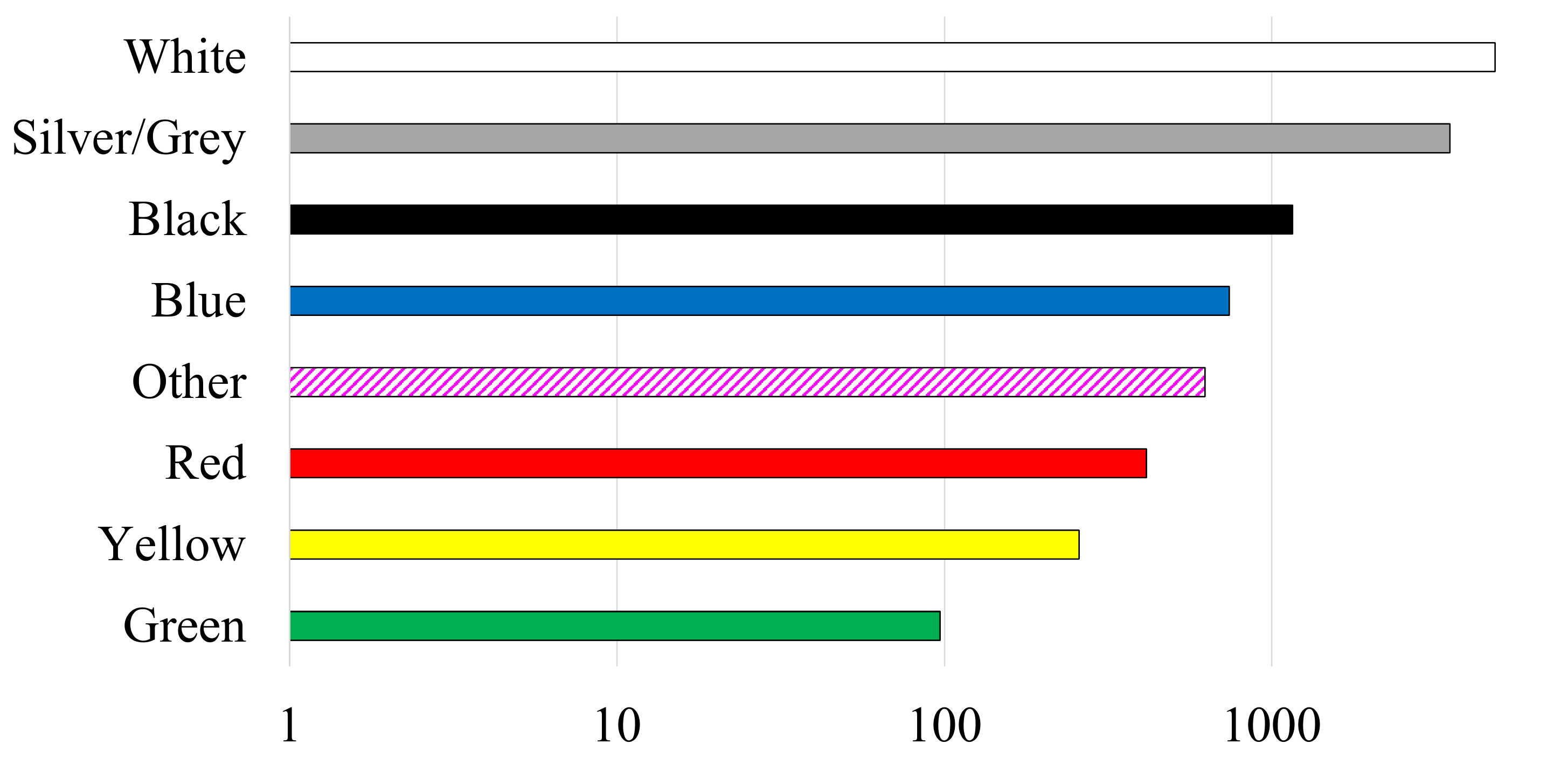}
  \caption{Number of items in color classes}
  \label{fig:colors}
\end{subfigure}
\caption{Log distribution of number of items in each class}
\label{fig:test}
\end{figure}

COFGA contains 2,104 images captured in various land types—urban areas, rural areas and open spaces—on different dates and at different times of the day (all performed in daylight). The images were taken with a camera designed for high-resolution vertical and oblique aerial photography mounted on an aircraft. Images also differ in the size of the covered land area, weather conditions, photographic angle, and lighting conditions (light and shade). In total, it contains 14,256 tagged vehicles, classified into four granularity levels:

\begin{itemize}
    
\item \textbf{Class—}The category contains only two instances: ‘large vehicle’ and ‘small vehicle’, according to the vehicle’s measurements.
\item \textbf{Subclass—}'Small vehicles' and 'large vehicles' are divided according to their kind or designation. 'Small vehicles' are divided into a 'sedan', 'hatchback', 'minivan', 'van', 'pickup truck', 'jeep', and 'public vehicle'. 'Large vehicles' are divided into a 'truck', 'light truck', 'cement mixer', 'dedicated agricultural vehicle', 'crane truck', 'prime mover', 'tanker', 'bus', and 'minibus'  (Figs \ref{fig:challenge-labels:subclass}, \ref{fig:vehicles}).
\item \textbf{Features—}This category addresses the identification of each vehicle’s unique features. The features tagged in the 'small vehicle' category are 'sunroof', 'luggage carrier', 'open cargo area', 'enclosed cab', 'wrecked' and 'spare wheel'. The features tagged in the 'large vehicle' category are 'open cargo area', 'AC vents', 'wrecked', 'enclosed box', 'enclosed cab', 'ladder', 'flatbed', 'soft shell box' and 'harnessed to a cart' (Figs \ref{fig:challenge-labels:features}, \ref{fig:features}).
\item \textbf{Object perceived color—}Identification of the vehicle’s color as would be used by the human analyst to describe the vehicle: 'white', 'silver/gray', 'black', 'blue', 'red', 'yellow', 'green', and 'other' (Figs \ref{fig:challenge-labels:color}, \ref{fig:colors}).
\end{itemize}
It should be noted that an object can be assigned with more than one feature but is assigned to exactly one class, one subclass and one color.

\subsection{Annotation Procedure}
\subsubsection{Phase 1: Initial Labeling}
Two independent aerial imagery analysis teams first systematically scanned each image and annotated detectable vehicles within a 4-point oriented BB. Each BB was labeled as either ‘small vehicle’ or ‘large vehicle’. Objects that appeared in more than one image were labeled separately in both images, but such cases were scarce. The BB were drawn on a local vector layer of each image; thus, their metadata entail local pixel coordinates and do not entail geographic coordinates. The quality of these initial detections and their matching labels was validated at the fine-grained labeling stage by aerial imagery analysis experts who corrected annotation mistakes and solved inconsistencies. 
\subsubsection{Phase 2: Fine-Grained Features Labeling}
At this point, a team of aerial imagery analysis experts systematically performed a fine-grained classification of every annotated object. To improve the efficiency of this stage, a web application was developed to enable a sequential presentation of the labeled BB and the relevant image crop (it also enabled basic analysis manipulations such as zooming and rotation of the image). For each BB, an empty metadata card was displayed for the analysts to add the fine-grained labels. After completing the second stage of labeling, the data was double-checked and validated by independent aerial imagery experts.

\subsection{Dataset Statistics}
In this section, we present the statistical properties of the COFGA dataset.

\subsubsection{Inter- and Intra-Subclass Correlation}
\begin{figure}
\centering
\includegraphics[width=\textwidth]{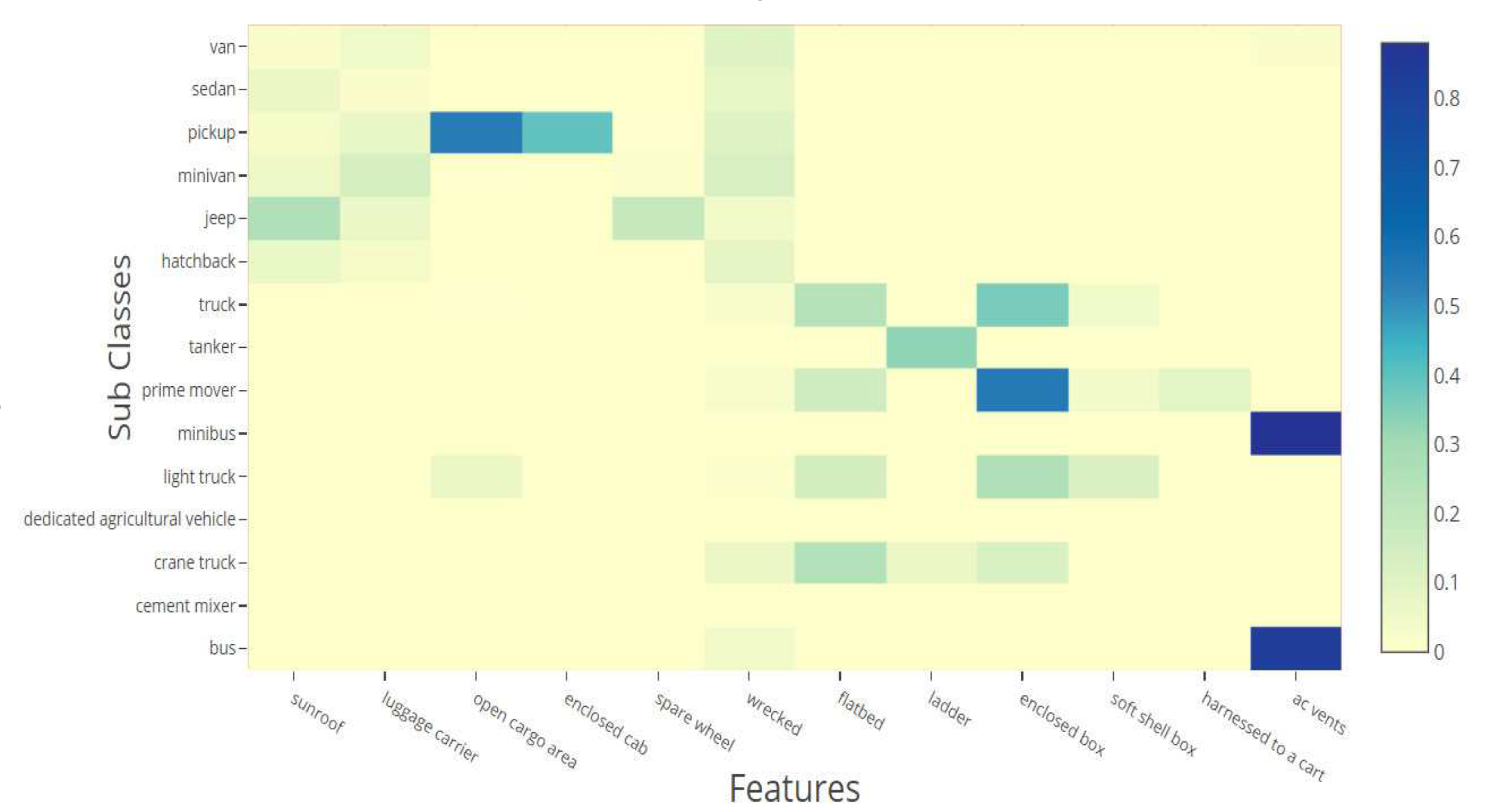}
\caption{Heat map of the inter- and intra-subclass correlation}
\label{fig:heatmap}
\end{figure}

Each object in the dataset is assigned to a multilabel vector. These labels are not independent; for example, most of the objects with a 'spare wheel' label belong to the subclass of 'jeep'. Fig \ref{fig:heatmap} shows the inter-subclass and intra-subclass correlation. The inter-subclass correlation is a measure of the correlation between different subclasses. From the features point of view, it is the measure of how the same feature is distributed in different subclasses. Hence, while exploring the heat map (Fig \ref{fig:heatmap}), the inter-subclass correlation can be seen in the value distributed in the columns of the heat map, where a high value means that the distribution of this feature has a peak for the specific subclass. The intra-subclass correlation is a measure of the correlation between different features for a specific subclass. Hence, while exploring the heat map (Fig \ref{fig:heatmap}), the intra-subclass correlation can be seen in the values of the rows of the heat map. One can see that the most common feature (in being shared through different subclasses) is the feature of the vehicle that is 'wrecked'. On the other hand, it can be seen that the most correlative subclass (in having the most significant number of different features) is a 'pickup'. Additionally, the most correlative pair of feature–subclass is a 'minibus' with 'air condition vents'.

\subsubsection{Pixel Area of Objects from Different Subclasses}
\begin{figure}
\centering
\includegraphics[width=100mm]{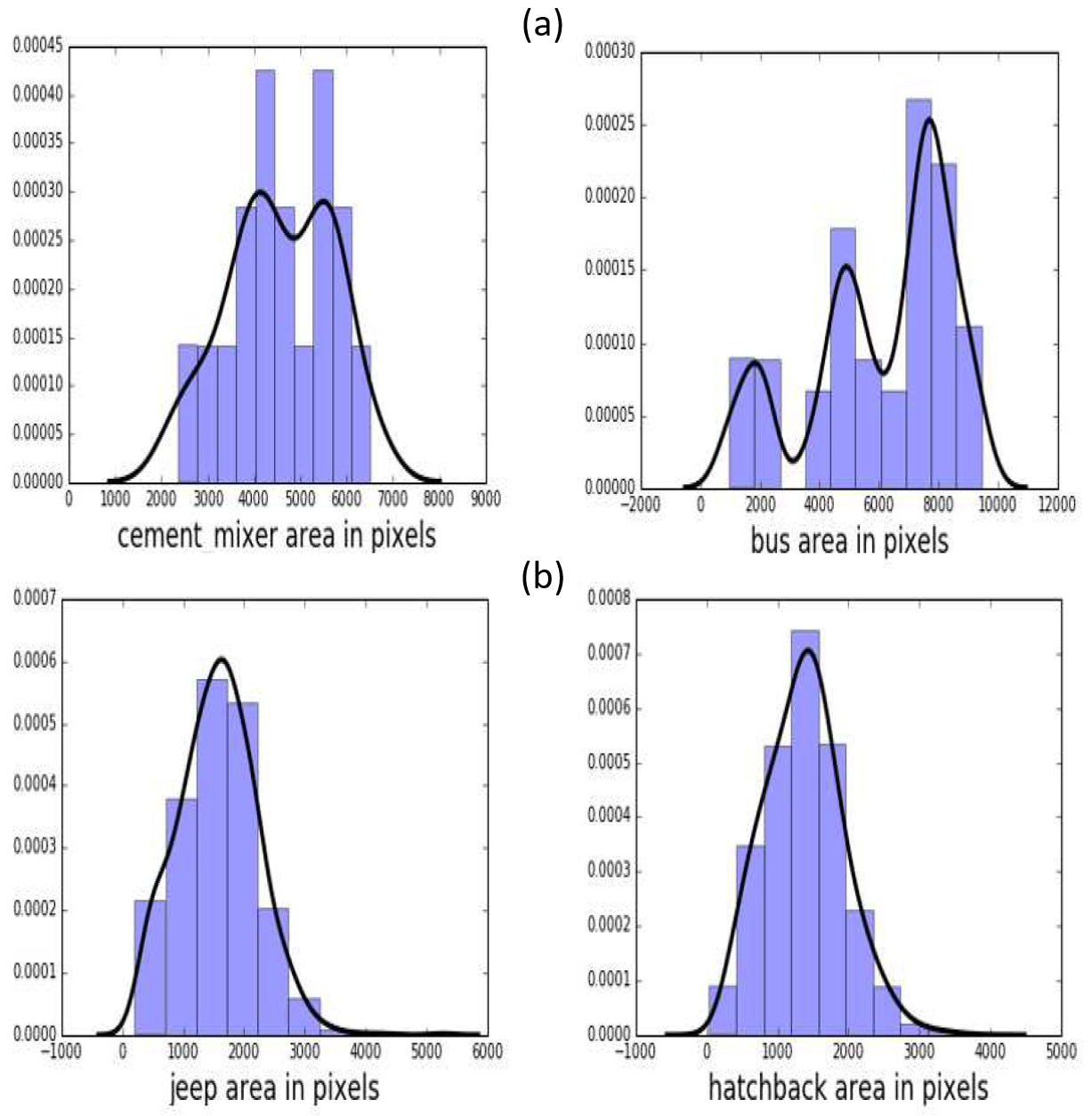}
\label{fig:aaa}
\caption{Distribution of the area, in pixels, of objects from different subclasses: (a) two subclasses of the 'large vehicle' class, (b) two subclasses of the 'small vehicle' class }
\label{fig:area}
\end{figure}

We compute the distribution of the areas, in pixels, of objects that belong to different subclasses, as shown in  Fig \ref{fig:area}. As explained above, the objects were annotated using images with a spatial resolution of 5–15 cm GSD. While exploring the distribution, one can notice that for a 'large vehicle' (Fig \ref{fig:area} a), there are usually 2–3 distinct peaks, while for a 'small vehicle' (Fig \ref{fig:area} b), the three peaks coalesce to one. Additionally, it is more common to find 'large vehicles' of different sizes than it is to find 'small vehicles' of different sizes, even when 'large vehicles' are generated from the same subclass (such as 'buses' of different sizes compared to 'hatchbacks' of different sizes). One can use this distribution to preprocess the images to better fit and train the classifiers for the task. One can also try to measure the GSD per image and cluster images by their GSD to better fit the potential pixel area of the specific classifier.

\section{Comparing Datasets}\label{sec:comparison}
In this section, we compare the datasets mentioned in section-\ref{sec:related} with the COFGA dataset.

xView \citep{xview} is the overhead imagery dataset with the largest number of tagged objects. It has more categories than COFGA, but its spatial resolution is 2–6 times poorer and its labeling ontology includes half of COFGA's granularity levels. 

DOTA and iSAID \citep{dota,isaid} also have more tagged objects than COFGA, but have less than half the number of categories, 1–3 times poorer spatial resolution, and a quarter of the granularity levels. 

COWC \citep{cowc} has more tagged objects than COFGA, but these annotations are merely CPM, which are unlike the oriented BB annotations in COFGA. COWC also has only a single category and 1–3 times poorer spatial resolution.

VisDrone \citep{visdrone} contains some aerial imagery, but most of the images are taken from a height of 3–5 m and are not in the same category as the other datasets mentioned here. For that reason, we decided it was worth mentioning but not suitable for comparison.


A comparison can be found in Table \ref{tabular_datasets}.

\begin{table}[htbp]
\centering

\vbadness=10000
\begin{tabular}{p{1.4cm}|p{1.8cm}p{1.6cm}p{1.5cm}p{0.9cm}p{0.7cm}p{1.4cm}p{1.4cm}}
\hline
\multicolumn{4}{r}{Dataset details}\\
\hline
\\Dataset& Annotation method& \#Categories& \#Granularity levels&\#Images & GSD (cm)& \#Object instances & \#Vehicle instances\\
\hline
DOTA 1.5 & \makecell{oriented\\BB}&  \makecell{16}&  \makecell{1}&  \makecell{2,806}&  \makecell{15}&  \makecell{471,438}&  \makecell{\textasciitilde 380K}\\
xView &\makecell{horizontal\\ BB}&  \makecell{60}&  \makecell{2}&  \makecell{1,127}&  \makecell{30}&  \makecell{\textasciitilde 1M}&  \makecell{\textasciitilde 280K}\\
COWC & \makecell{centroid pixel\\map}&  \makecell{1}&  \makecell{1}&  \makecell{2418}&  \makecell{15}&  \makecell{37,716}&  \makecell{37,716}\\
iSAID & \makecell{polygon\\segmentation} & \makecell{15}& \makecell{1}& \makecell{2,806}& \makecell{15}& \makecell{655,451}& \makecell{\textasciitilde 380K}\\
\hline
\textbf{COFGA}&  \makecell{oriented\\BB}& \makecell{37}& \makecell{\textbf{4}}& \makecell{2,104}& \makecell{\textbf{5-15}}& \makecell{14,256}& \makecell{14,256}\\
\hline

\end{tabular}
\captionof{table}{Detailed dataset comparison}
\label{tabular_datasets}

\end{table}

\section{MAFAT Challenge}
In this section, we briefly describe the competition and a few notable solutions used to tackle the challenge of fine-grained classification.
 
 Because detection in aerial images has become a relatively easy task with the help of dense detectors such as RetinaNet \citep{retinanet},
 the competition was focused on fine-grained classification. The BBs of the vehicles were given to the participants, who were asked to train fine-grained classification models. 

In addition to the labeled training set, participants received an unlabeled test set, constructed using three subsets:
\begin{enumerate}
    \item \textbf{A public test set} that included 896 objects. This test set was used to calculate the score shown on the public leaderboard during the public phase of the competition.
    \item \textbf{A private test set} that included 1,743 objects. This test set was used to calculate the final score, shown on the private leaderboard that was used for final judging.
    \item \textbf{Noise,} which included 9,240 objects and was automatically tagged by the detection algorithm we used on the images in the test set. These objects were not tagged by us and were not used to compute the submission score. We added these objects as distractors to reduce the likelihood of a competitor winning this challenge by cheating through manually tagging the test set.
\end{enumerate}

\subsection{Evaluation Metric}
For each label (where 'label' represents all distinct classes, subclasses, unique features and perceived colors), an average precision (AP) index is calculated separately (Equation \ref{eqn:AP}). 
\begin{equation}\label{eqn:AP}
AP(label) = \frac{1}{K}\sum_{k=1}^{n}Precision(k)rel(k)
\end{equation}
where $K$ is the number of objects in the test data with this specific label, $n$ is the total number of objects in the test data, $Precision(k)$ is the precision calculated over the first k objects, $rel(k)$ equals to 1 if the object-label  prediction for object k is $True$ and 0 if it is $False$.
\newline
Then, a final quality index—mean average precision (mAP)—is calculated as the average of all AP indices.  (Equation \ref{eqn:mAP}).
\begin{equation}\label{eqn:mAP}
mAP = \frac{1}{Nc}\sum_{label=1}^{N_c}AP(label)
\end{equation}
where $Nc$ is the total number of labels in the dataset.

This index varies between 0 and 1 and emphasizes correct classifications with significance to confidence in each classification, aiming to distinguish between participants who classify objects correctly, in all environmental conditions, as well as to reference their confidence in the classification.

The weighting to the left of Equation 2 ensures that all labels in the dataset have equal influence on the mAP, regardless of their frequency. This is a focal detail due to the large variance in the frequency of the various labels (classes, subclasses, unique features and perceived colors). As a result, a minor improvement in performance on a rare class can be equivalent to a major improvement on a more frequent class. Another noteworthy property of the mAP metric is that it does not directly assess the output of the model. Rather, it measures only the ordering among the predictions without taking the actual prediction probabilities into account.

These properties make the mAP evaluation metric suitable for the multilabel-multiclass problem with highly imbalanced data.

\subsection{Challenging Aspects of the Competition}
\label{sec:challange}

The most challenging aspects of the competition were:

\textbf{Rare Labels—}Some subclasses and features contained a small number of objects in the training set (with a minimum of 3 objects for the ladder feature). Modern computer vision models, particularly convolutional neural networks (CNNs), are very powerful due to their ability to learn complex inputs to output mapping functions that contain millions and sometimes billions of parameters. On the other hand, those models usually require large amounts of labeled training data, usually on the scale of thousands of labeled training examples per class. Training CNNs to predict rare labels is a complex challenge, and one of the competition goals was to explore the creative methods of doing so with a small amount of labeled data.

\textbf{Imbalanced Data—}Not all of the labels were rare. Some classes and subclasses in the training set had thousands of labeled examples (Fig \ref{fig:dist}). This imbalanced nature of the training set presented a challenge to the participants but also valued the attractive potential of transfer learning and self-supervised learning methods. We were interested in exploring which techniques are effective in tackling imbalanced classes.

\textbf{Train–Validation–Test Split—}Participants received only the labeled training set and the unlabeled test set. The test set was split into public and private test sets; however, the characteristics of the split were not exposed to the participants. Learning the parameters of a prediction function and testing it on the same data is a methodological mistake that causes overfitting. In the case of imbalanced data and some infrequent classes, the challenge of splitting the data into training, validation, and test sets is enhanced.

\textbf{Label Diversity—}The labels were diverse. Some were color-based (\emph{perceived color}), some were based on shapes (i.e., \emph{tanker}), some were based on many pixels (large objects, i.e., \emph{semitrailer}), some were based on a small number of pixels (fine-grained features, i.e., \emph{sunroof}, \emph{ladder}), and finally, some required spatial context (i.e., a \emph{wrecked car}, which is very likely to be surrounded by other wrecked cars). 

\subsection{Notable Solutions}\label{sec:part}
We describe the details of the winning solution and a few notable approaches used by the participants to tackle the challenge, including one baseline solution published by the MAFAT Challenge team.

\subsubsection{Baseline Model (mAP: 0.60)}
To evaluate the challenge’s difficulty, MAFAT also released results for fine-tuning the existing state-of-the-art object classification models using a simple rotation augmentation. The details of this baseline research were published before the competition, allowing participants to spend their time on creative approaches and improvements.

MAFAT compared the results of two different architectures—MobileNet \citep{mobilenet} and ResNet50 \citep{resnet}—using different loss functions—cross entropy and weighted cross entropy—and either using or not using rotation augmentations. The best baseline model was based on MobileNet \citep{mobilenet}, which used both the rotation augmentation and a weighted cross entropy loss and achieved a 0.6 mAP score.

\begin{figure}[ht]
\centering
\includegraphics[width=150mm]{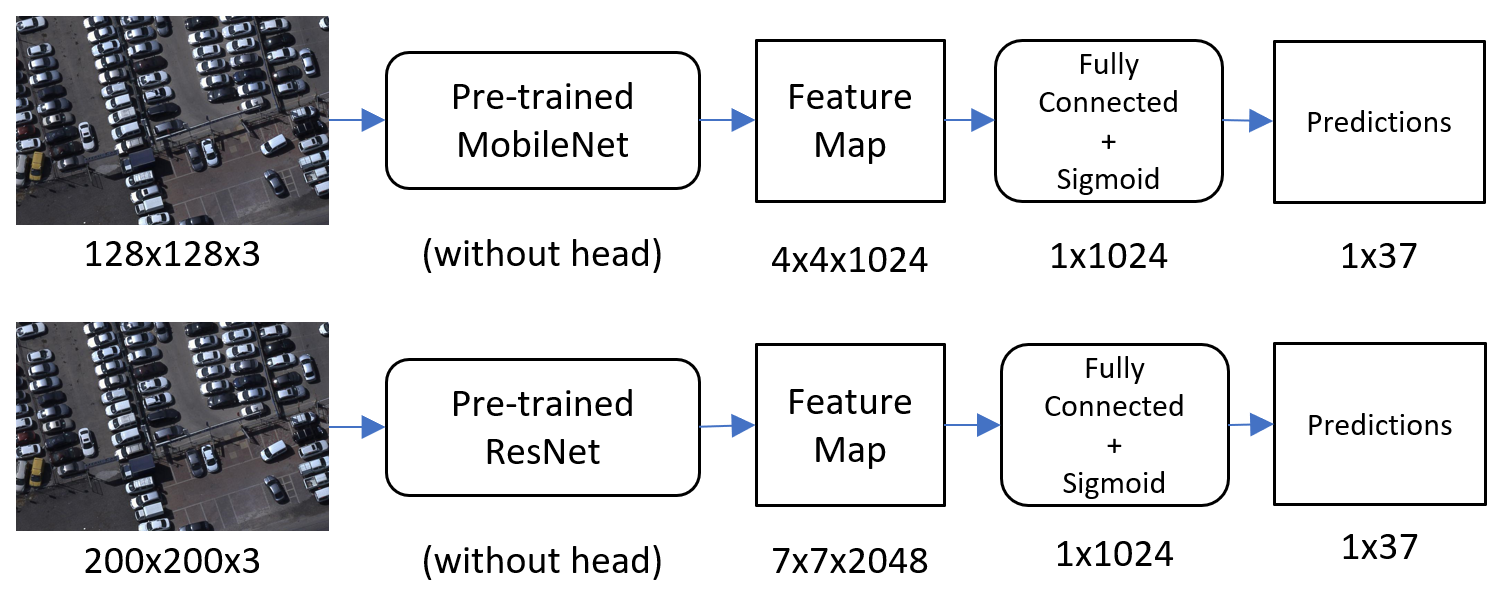}
\caption{Architectures used in the baseline: based on MobileNet and ResNet50}
\label{fig:baseline-architecture}
\end{figure}

\subsubsection{1st Place—SeffiCo—Team MMM (mAP: 0.6271)}
The MMM team used an aggressive ensemble approach, based on  the following sequence:
\begin{enumerate}
    \item \textbf{Train a very large number of different models} (hundreds) by using as many differentiating mechanisms as possible, such as bagging, boosting, different architectures, different preprocessing techniques, and different augmentation techniques.
    \item \textbf{Cluster} the trained models into different groups by using the k-means clustering algorithm on their predictions. This stage aims to generate dozens of model clusters in such a way that the models in the same cluster are more similar to each other than to those in the other clusters, ensuring diversity between the clusters.
    \item \textbf{Pick} the best representative models from each cluster.
    \item \textbf{Ensemble} the representative models into an integrated model by averaging their logits.
\end{enumerate}

\textbf{Preprocessing—}
\begin{enumerate}
    \item \textbf{Cropping—}The first step is to crop the vehicles from the image using the coordinates from the provided input data, but after observing the data, it was noticeable that the area around the vehicles could reveal information about the vehicles’ properties (spatial context). Each vehicle was cropped with an additional padding of five pixels from the surrounding area (Fig \ref{fig:rotation}).
    \item \textbf{Rotation—}The vehicles in the images are not oriented in any single direction, which adds complexity to the problem. To solve this challenge, the model uses the fact that the length of a vehicle is longer than its width, and this fact is used to rotate the cropped image and align the larger edge with the horizontal axis.
\begin{figure}[ht]
\centering
\includegraphics[width=120mm]{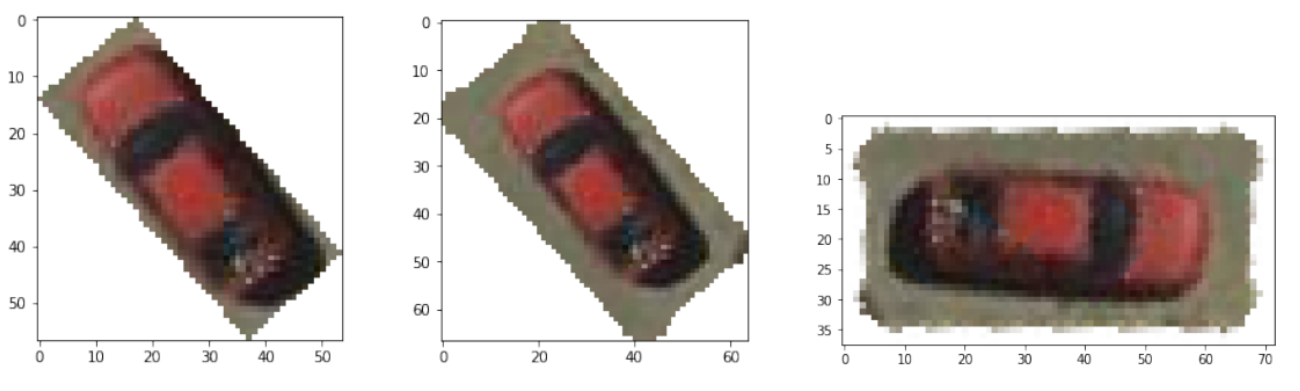}
\caption{Padding, cropping and rotation}
\label{fig:rotation}
\end{figure}    
    \item \textbf{Color Augmentation—}Converting the images from RGB (red-green-blue channels) to HSV (hue-saturation-value) and swapping color channels are methods used to gather more color data (Fig \ref{fig:color-channels}).
    \item \textbf{Squaring—}Since most existing CNN architectures require a fixed-sized square image as input; hence, an additional zero padding is added to each cropped image on the narrower side, and then the image is resized to fit the specific size of each architecture (79, 124, 139 and 224 pixels), similar to \citep{segmentation}.
\begin{figure}[ht]
\centering
\includegraphics[width=100mm]{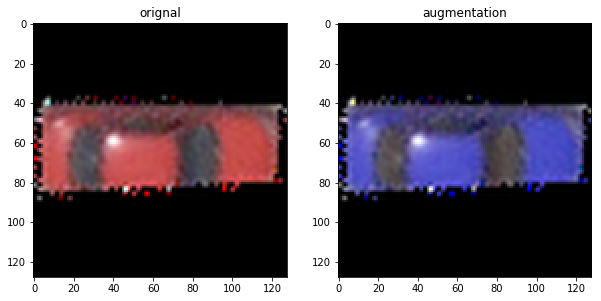}
\caption{Squaring and color augmentation: obtained by permuting the 3 channels of the RGB image}
\label{fig:color-channels}
\end{figure}
    \item \textbf{Online—}Finally, online data augmentation is performed in the training process using the Imgaug library.
\end{enumerate}

After handling the data in different ways, team MMM trained hundreds of different models, including almost all of the state-of-the-art classification models, such as Inception V3 \citep{inception}, Xception \citep{xception}, NASNet \citep{nasnet}, and a few versions of ResNet \citep{resnet} to predict all 37 labels (classes, subclasses, unique features and perceived colors).

They created a unique ensemble for each label (or group of similar labels). This means that they could choose the best subset of the different models for one specific label, e.g., determining whether the vehicle has a 'sunroof'. Once they had the logits from each model, they combined them using a method, which they engineered to work well with mAP, that transforms the predicted probabilities into their corresponding ranks. Then the mean between the ranks of the different models was computed. This method allowed the authors to average the ranks of the predictions rather than average the probabilities of the predictions. 
\begin{figure*}
\centering
\includegraphics[width=150mm]{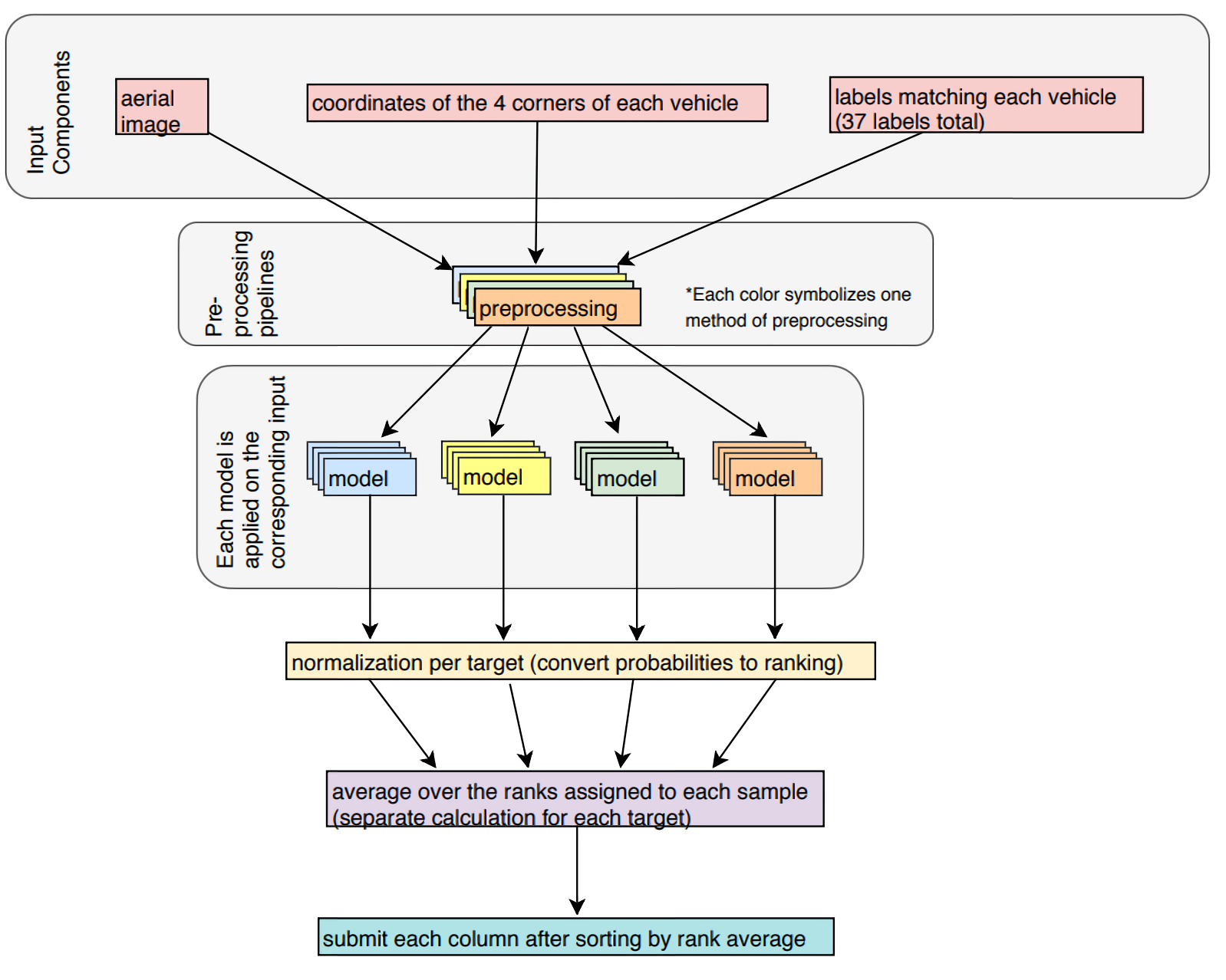}
\caption{Diagram of the ensemble pipeline used by SeffiCo-Team MMM}
\label{fig:seffi-model}
\end{figure*}

In their submissions, the authors’ main approach was to keep modifying the different stages of the pipeline to generate more models. For each pipeline, they saved the predictions of the new model and the model itself. They ultimately obtained more than 400 sets of models, predictions and their corresponding public mAP scores, evaluated against the public test set. Each model had a unique combination of preprocessing, augmentation, and model architecture (Fig \ref{fig:seffi-model}).

\subsubsection{Yonatan Wischnitzer (mAP: 0.5984)}
This participant exploited the hierarchical nature of the labels. In other words, given a known label, the distribution of the other labels for the same object changes, and therefore, the participant used a step-by-step approach.

\begin{figure}[ht]
\centering
\includegraphics[width=120mm]{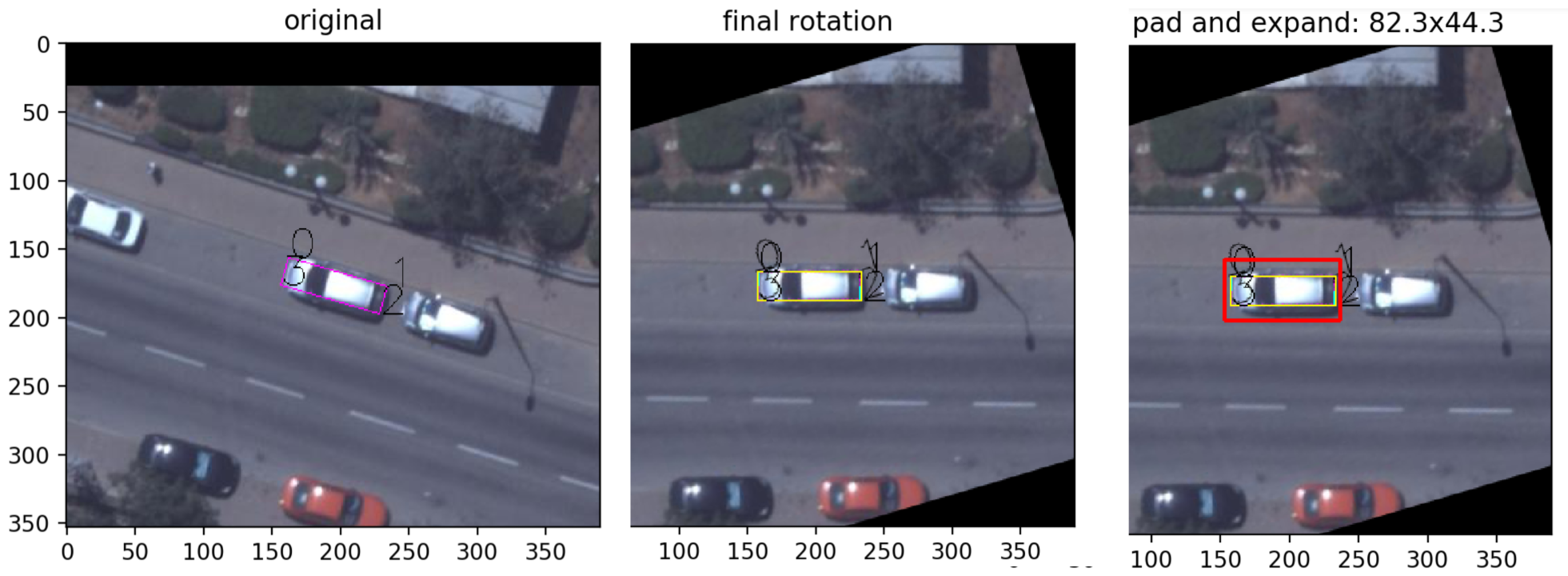}
\caption{Yonatan Wischnitzer—preprocessing}
\label{fig:akavish-input-rotate}
\end{figure}

For preprocessing, the participant first performed an alignment rotation (parallel to axes) and then padded the objects by a 2:1 ratio scaled to 128 × 64 pixels (Fig \ref{fig:akavish-input-rotate}). This allowed the participant to reduce the complexity of the structure of the network, as all the objects were approximately horizontally aligned. Furthermore, extra auxiliary features with the size of the object in the source image were added to each object; 
each object was normalized by the average size of all objects in that same image and by the source file type—JPG or TIFF. For online data augmentation, the participant used a sheer effect to mimic the different image capturing angles.

\begin{figure}[ht]
\centering
\includegraphics[width=120mm]{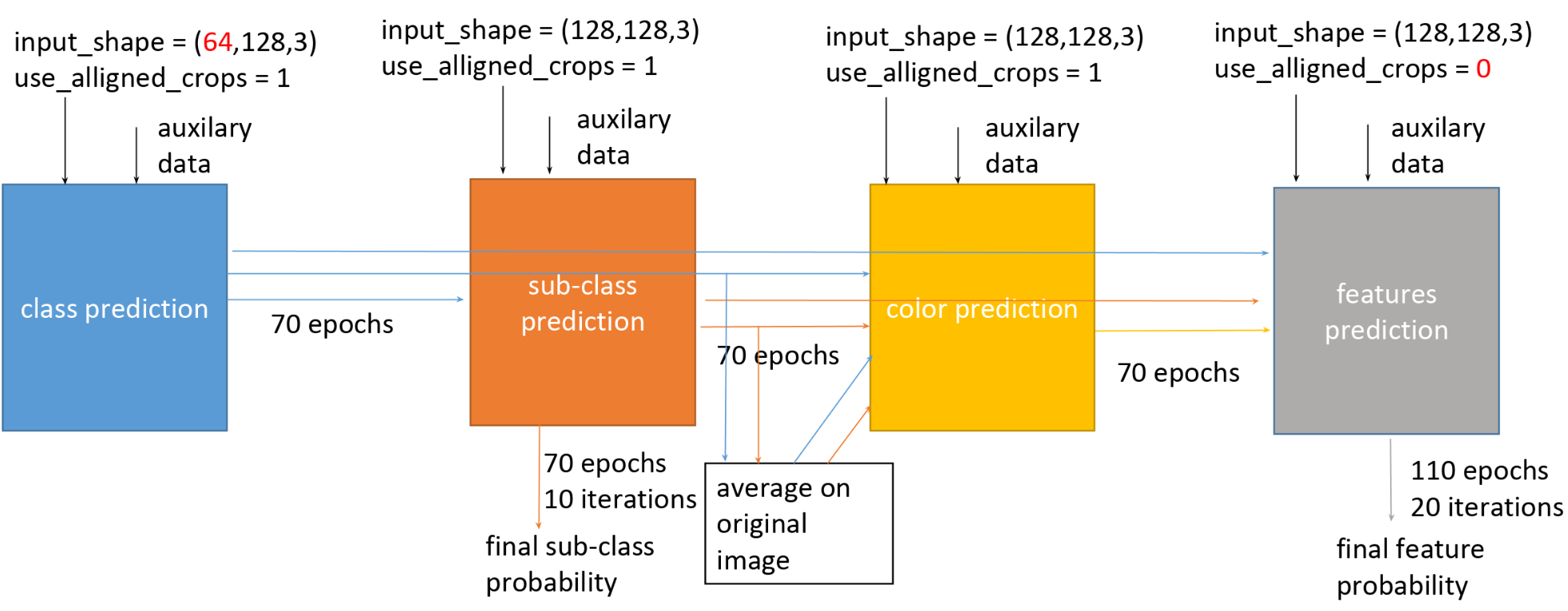}
\caption{Yonatan Wischnitzer's model architecture, exploiting the hierarchical nature of COFGA dataset's tagging taxonomy}
\label{fig:akavish-hirarchy}
\end{figure}

The participant’s approach was to use the entire training set for training and to use the public test set for validation using the submission system. In total, four models were trained, each predicting a subset of labels based on the knowledge of some of the other labels. The first model predicts the class from the image patch, the second model predicts the subclass from the image patch and uses the class predicted by the first model as auxiliary data, and the third model predicts the color and takes the class and subclass predicted by the first and second models as auxiliary data as well as the average RGB channels from the original image patch. Finally, the fourth model predicts the remaining features and takes the class, subclass, and color, predicted by the three previous models, as auxiliary data (Fig \ref{fig:akavish-hirarchy}). Each model is trained with the crossentropy loss from each category and is based on a fine-tuned MobileNet \citep{mobilenet} architecture, as it has a significantly small number of parameters in comparison to other state-of-the-art classification networks.

\section{Conclusions}\label{sec:conclusions}
In this paper, we introduced COFGA: a dataset for fine-grained classification of objects from aerial images, and compared it to leading public overhead imagery datasets. We showed that although the number of objects in the COFGA dataset is relatively small, it has a spatial resolution that is 1-6 times better than that of the other leading pubic datasets, and its granularity is 1-4 times better than all the granularities of the datasets mentioned in section \ref{sec:related}. With the exception of the xView dataset \citep{xview}, COFGA has more than double the number of categories. These qualities make COFGA a great dataset for overhead imagery analysis tasks, and specifically to fine-grained classification.
We also introduced the MAFAT Challenge, in which the participants were asked to address the classification of fine-grained objects in COFGA's images, which is a multiclass, multilabel problem. The main challenges were imbalanced data and low-resource training (\S\ref{sec:challange}). These challenges were addressed with different strategies, including aggressive-ensemble techniques, data augmentation, data preprocessing, and hierarchical model assembly. We described the notable proposed solutions (\S\ref{sec:part}).

We have offered the COFGA dataset for public use and hope that this dataset, along with the competition for fine-grained classification, will give rise to the development of new algorithms and techniques. We believe that the high-resolution images and the accurate and fine-grained annotations of our dataset can be used for the development of technologies for automatic fine-grained analysis and labeling of aerial imagery.

\bibliography{references}

\end{document}